%% file: main.tex
\documentclass[12pt]{article}
\usepackage[utf8]{inputenc}
\usepackage{graphicx}
\graphicspath{Figures/}
\usepackage{amsmath}
\usepackage{amssymb}
\usepackage{listings}
\usepackage{hyperref}
\usepackage{cuted}
\usepackage{authblk}
\usepackage{threeparttable}
\usepackage{tikz}
\usepackage{setspace}
\usepackage{float}
\usepackage{lineno}
\usepackage{algorithm}
\usepackage{algpseudocode}
\usepackage{xspace}
\usepackage{colortbl}
\usepackage{breqn}

\newcommand{\floatfootnote}[1]{\ifx\[$\else\footnote{#1}\fi}
\usepackage[authoryear]{natbib}

\usepackage{bbm}
\input{commands}

\title{\vspace{-4em} Amortized Bayesian Local Interpolation NetworK: Fast covariance parameter estimation for Gaussian Processes}
\author[1]{Brandon R. Feng}
\author[2]{Reetam Majumder}
\author[1]{Brian J. Reich}
\author[3]{Mohamed A. Abba}
\affil[1]{Department of Statistics, North Carolina State University}
\affil[2]{Department of Mathematical Sciences, University of Arkansas}
\affil[3]{Amazon Research}

\begin{document} 

\maketitle
\vspace{-2em}
\begin{abstract}\begin{singlespace}\noindent
  Gaussian processes (GPs) are a ubiquitous tool for geostatistical modeling with high levels of flexibility and interpretability, and the ability to make predictions at unseen spatial locations through a process called Kriging. Estimation of Kriging weights relies on the inversion of the process' covariance matrix, creating a computational bottleneck for large spatial datasets. In this paper, we propose an Amortized Bayesian Local Interpolation NetworK (A-BLINK) for fast covariance parameter estimation, which uses two pre-trained deep neural networks to learn a mapping from spatial location coordinates and covariance function parameters to Kriging weights and the spatial variance, respectively. The fast prediction time of these networks allows us to bypass the matrix inversion step, creating large computational speedups over competing methods in both frequentist and Bayesian settings, and also provides full posterior inference and predictions using Markov chain Monte Carlo sampling methods. We show significant increases in computational efficiency over comparable scalable GP methodology in an extensive simulation study with lower parameter estimation error. The efficacy of our approach is also demonstrated using a temperature dataset of US climate normals for 1991--2020 based on over 7,000 weather stations.  \vspace{12pt}\\

 \end{singlespace}\end{abstract}

\section{Introduction} \label{s:intro}

Gaussian Processes (GPs) are a highly powerful and flexible class of models that have been a popular framework commonly utilized in machine learning, and more specifically, the field of spatial statistics \citep{rasmussen2003gaussian, banerjee2008gaussian, gelfand2016spatial}. Statistical analysis of spatial data often assumes that there exists a decaying correlation structure along locations in a certain domain. GPs allow the user to specify a parameterized spatial covariance function in the response a priori. Given the observed data, this allows both parameter estimation and predictions at unobserved locations through a spatial interpolation process called 
Kriging \citep{cressie1990origins}. However, Kriging requires inverting the covariance matrix during evaluation of the joint likelihood function, which causes runtime to be cubic in the number of locations and quadratic in memory. This computational cost is prohibitive for analyzing large spatial datasets, and has spurred research into methods for scalable GP inference. 

Sparse approximations of the likelihood are a popular class of methods striving to make GPs scalable. This includes covariance and precision matrix tapering \citep{furrer2006covariance, kaufman2008covariance, rue2009approximate}, which enforces a specific sparse structure on the matrix, and the Vecchia approximation \citep{vecchia1988estimation,stein2004approximating}. The Vecchia approximation splits the full likelihood into a product of univariate conditional likelihoods, each of which is simplified to condition on a small subset of ordered neighbors. This process gives the precision matrix a sparse block diagonal form, significantly reducing the computational cost of likelihood evaluation. \cite{datta2016hierarchical} built upon this concept of conditioning on ordered neighbor subsets with the Bayesian nearest-neighbor Gaussian process (NNGP) methodology. The Vecchia approximation is shown to be a close approximation of the true GP in terms of Kullback-Leibler divergence \citep{guinness2018permutation} under certain grouping methods, however, it can falter in cases of high noise \citep{katzfuss2021general} and cause inaccuracies in parameter estimation. 

Recent advances in deep learning has led to an alternative way of scalable spatial analysis via directly estimating model parameters using pre-trainned neural networks. \cite{gerber2021fast} pre-trained neural networks on tens of thousands of fields of data, each simulated from a GP with a different parameter combination, to predict GP parameters. \cite{lenzi2023neural} proposed pre-training neural networks with data simulated from plausible parameter ranges to predict the process parameters of a spatial extremes model. \cite{sainsbury2023neural,sainsbury2024likelihood} used neural architectures to directly estimate GP and max-stable process parameters from observed data. \cite{majumder2023deep,majumder2024modeling} utilized deep learning to approximate intractable joint densities for spatial extremes, embedding a surrogate likelihood within a Bayesian framework. Similarly, \cite{walchessen2024neural} pre-trained neural networks on simulated data to approximate the likelihood function, replacing the computationally expensive likelihoods with the network for maximum likelihood estimates of parameters in Gaussian and Brown-Resnick processes. 

These approaches can be considered a form of amortized learning, based on first pre-training neural networks on simulated data from various parameter combinations for a statistical model that is challenging to analyze using classical tools, and then using the fitted neural network to predict the parameters or the likelihood in a different dataset. There is an initial overhead cost stemming from data generation and pre-training; however, this amortizes the computational cost of conventional parameter estimation since the same pre-trained network can be used to estimate parameters of multiple datasets. However, most of these models come with caveats that limit their wider usability. Several have been developed for gridded data in an attempt to leverage the predictive power of convolutional neural networks, and do not translate well to irregularly observed data. They are also overwhelmingly frequentist approaches, limiting uncertainty quantification to bootstrap methods. Finally, some of the models which are designed to provide inference in irregularly spaced data nevertheless require the entire model to be re-fit in order to make predictions at new locations.
A GP with its computational bottlenecks amortized using deep learning should be able to provide posterior inference on parameters as well as predictions at unobserved locations in a Bayesian framework, for large, irregularly spaced data, in a tractable manner. These desirable properties of a GP motivate this work.

\paragraph{Our contribution:} We propose the Amortized Bayesian Local Interpolation NetworK, or A-BLINK, which uses neural networks to approximate the Kriging weights and spatial variance used to evaluate the likelihood of a Vecchia approximated GP. 
Predictions from the neural network can be used to evaluate a surrogate likelihood for the GP, bypassing the need for expensive matrix inversions. 
To the best of our knowledge, this is the first approach to use a single set of pre-trained neural networks to do full Bayesian posterior inference and provide predictions at unobserved locations for arbitrary, ungridded spatial datasets of 10,000 locations or higher. Through simulated and real-world data, we demonstrate that A-BLINK is competitive with state-of-the-art scalable GP algorithms for spatial data in terms of parameter estimation while providing significant computational speedups.

\section{Gaussian Process Methodology} \label{s:GP_Method}

Let \textbf{Z} be an isotropic GP over a spatial domain $\calD\subset\mathbb{R}^2$. At $n$ locations denoted $\{\bs_1,...,\bs_n\}$ where $\bs_i = (s_{i1},s_{i2}) \text{ for } i \in \{1,...,n\}$, we observe realizations of \textbf{Z}, $Z_i = Z(\bs_i)$. We assume \textbf{Z} has mean $\mbox{E}(Z_i) = \mu$ and spatial variance $\mbox{Var}(Z_i) = \sigma^2$. Additionally, let the spatial correlation between a pair of locations be $\text{Cor}\{Z(s_i), Z(s_j)\} = r\cdot K(d_{ij})$, where $r \in [0,1]$ is the proportion of spatial variance, $K$ is a valid correlation kernel, and distance $d_{ij} = ||\bs_i - \bs_j||$ for locations $i, j \in \{1,...,n\}$. We use the $\Matern$ correlation kernel \citep{matern1960spatial, stein2012interpolation} given by:
\begin{equation} \label{eq:Matern Corr}
    K(d) = \frac{1}{\Gamma(\nu)2^{\nu-1}}\left(\frac{d}{\phi} \right)^{\nu} \mathcal{K}_{\nu}\left(\frac{d}{\phi} \right),
\end{equation}

\noindent where spatial range $\phi > 0$, smoothness $\nu > 0$ and $\mathcal{K}_{\nu}$ is the modified Bessel function of the second kind. This gives GP parameter set $\boldsymbol{\theta} = (\phi, \nu, r)$ and  full paramter set $\Theta = \{\mu, \sigma^2, \boldsymbol{\theta}\}$.

The joint distribution of \textbf{Z} is multivariate normal with density
\begin{equation}\label{eq:Full_LogLike}
\footnotesize 
\begin{aligned}
    f(\mathbf{Z}|\theta) &\propto \det(\Sigma(\Theta))^{-1/2} \exp\left\{-\frac{1}{2}(\mathbf{Z} - \boldsymbol{\mu})^\top \Sigma(\Theta)^{-1} (\mathbf{Z} - \boldsymbol{\mu})\right\}, \\
    &\text{for covariance} \quad \Sigma(\Theta) = \sigma^2[r\mathbf{K} + (1 - r)\mathbf{I}_n],
\end{aligned}
\end{equation}


where $\mathbf{K}_{ij} = K(d_{ij})$. In Bayesian GP analysis, the posterior distributions stemming from this likelihood are computationally intractable to directly evaluate. As a result,  Markov chain Monte Carlo (MCMC) sampling is usually employed for posterior inference. Evaluation of the log-likelihood, 
\begin{equation}\label{eq:Log_LogLike}
\footnotesize 
 \ell(\bZ | \Theta) \propto  - \frac{1}{2}\log \det \Sigma(\Theta) - \frac{1}{2} (\bZ - \boldsymbol{\mu})^T\Sigma(\Theta)^{-1}(\bZ-\boldsymbol{\mu}),
\end{equation}
is computationally expensive since it involves the determinant and inverse of the $n\times n$ matrix $\Sigma(\Theta)$. This typically requires $O(n^3)$ operations, rendering parameter estimation and predictions at unobserved locations prohibitive for large spatial datasets. 

\subsection{The Vecchia approximation}

A common approach to reducing this cost is the Vecchia approximation \citep{vecchia1988estimation, stein2004approximating, katzfuss2021general}. The joint distribution of $\bZ$ can be written as the product of univariate conditional distributions:

\begin{equation}\label{eq:jointPDF1}
   f(\bZ|\Theta) = \prod_{i=1}^n f(Z_i|Z_1,...,Z_{i-1},\Theta).
\end{equation}

\begin{figure}[ht!]
    \centering
    \includegraphics[width=0.95\linewidth, scale = 1.5]{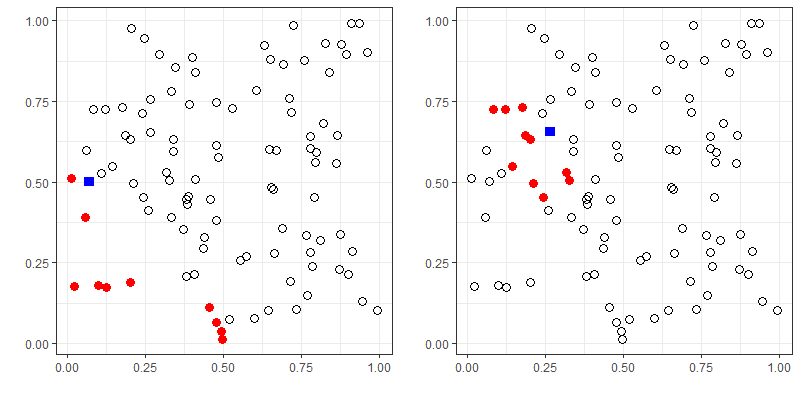}
    \caption{Illustration of neighbor sets (red) of size $m=10$ for each location (blue) in the Vecchia approximation, for a random set of $n = 100$ locations on the unit square.}
    \label{fig:VecchiaNeighbor}
 \end{figure}

\noindent The Vecchia approximation conditioning $Z_i$ on only a small subset of $\{Z_1,...,Z_{i-1}\}$ of size $m$. Let $\calN_i = \{i_{1},...,i_{m_i}\} \subseteq \{1,...,i-1\}$ represent the $m_i \leq m$ indices of the conditioning set for observation $i$ and $\bZ_{(i)} = (Z_{i_1},...,Z_{i_{m_i}})$ represent the GP values of the conditioning set. Without loss of generality, assume the conditioning set is arranged so its locations are in nondecreasing distance from $\bs_i$. A visual of conditioning sets for two points is shown in Figure \ref{fig:VecchiaNeighbor}. Note the selected conditioning set is not necessarily the nearest neighbors to the location. The Vecchia approximation can therefore be written as
\begin{equation}\label{eq:jointPDF2}
   f(\bZ|\Theta) = \prod_{i=1}^n f(Z_i|Z_1,...,Z_{i-1},\Theta) \approx \prod_{i=1}^n f(Z_i|\bZ_{(i)},\Theta).
\end{equation}

Since the conditional distributions of a multivariate normal distribution are normal, we can write them as Kriging equations:
\begin{equation}\label{eq:condfull}
\small
Z_i|\bZ_{(i)},\Theta \sim \mbox{Normal}\left(\mu+\sum_{j=1}^{m}w_{ij}(Z_{(i)_{j}}-\mu),\sigma^2\exp(w_{i0})\right)
\end{equation}

\noindent where $m = \min(m_i, i-1)$. The weights and variance terms are given by:
\begin{equation}\label{eq:kriging_eqn}
\begin{split}
    \mbox{Kriging weights: }& (w_{i1},...,w_{im})^T = \bSigma_{i(i)}\bSigma_{(i)(i)}^{-1} \mbox{\ , \ \ \ }\\
    \mbox{Variance: }& \exp(w_{i0}) = 1-\bSigma_{i(i)}\bSigma_{(i)(i)}^{-1}\bSigma_{i(i)}^T\mbox{\ , \ \ \ }
\end{split}
\end{equation}

where \noindent $\bSigma_{i(i)} = \mbox{Cov}(Z_i,\bZ_{(i)})$ is an $1\times m$ vector and $\bSigma_{(i)(i)} = \mbox{Cov}(\bZ_{(i)})$ is a $m\times m$ matrix. For notational convenience, we will use $m$ to denote $m_i$ for the remainder of the paper. This approximation leads to large computational savings when $m \ll n$ as it reduces the cost of likelihood evaluation to $O(nm^3)$.

The expression in \eqref{eq:condfull} is a valid likelihood that can be used in both frequentist and Bayesian inference. Details of the frequentist implementation can be found in the supplementary material. In a Bayesian setting, placing a prior on the parameters, $\pi(\Theta)$, and taking the log of \eqref{eq:jointPDF2}, $\ell(\Theta)$, gives the posterior $p(\Theta|\bZ)$:
\begin{equation}\label{eq:Vecchia_Post}
\begin{split}
\centering
    \ell(\Theta) &= \sum_{i=1}^n \log f(Z_i|\bZ_{(i)}, \Theta)\\
    \log p(\Theta|\bZ) &= \ell(\Theta) + \log \pi(\Theta).
\end{split}
\end{equation}

\noindent The cost of posterior evaluation is still $O(nm^3)$ as the cost of likelihood evaluation overtakes the cost of the prior. While the Vecchia approximation can significantly reduce the computational cost, for large $n$, this cost can be prohibitive, especially in Bayesian analysis where the likelihood must be evaluated at each MCMC iteration. 

Since the Kriging weights $(w_{i1},...,w_{im})^T$ and log-variance $w_{i0}$ are deterministic functions of the spatial configuration of the locations and of the parameters, we propose replacing these matrix inversion calculations with predictions from pre-trained neural networks. This reduces computation cost down to matrix multiplications, which are easily parallelized, and reduces computational overhead with inversions typically requiring multiple intermediate calculations.

\section{The A-BLINK Algorithm} \label{s:A-BLINK_Method}

In this section, we will first introduce the neural network training algorithm and then outline how it will be used to amortize the likelihood calculation. For this algorithm to be applicable to any ungridded dataset, we assume that all spatial locations over the domain $\mathcal{D}$ are scaled to lie in $\mathbb{R}^{[0,1] \times [0,1]}$. Scaling the domain in this manner allows us to constrain the range of the parameter $\phi$, which depends on the spatial dimension. We also assume the marginal mean and variance are 0 and 1, respectively, as any dataset can be centered and scaled to have $\mu = 0$ and $\sigma^2=1$.

\subsection{Network training}

From \eqref{eq:Matern Corr} and \eqref{eq:kriging_eqn}, we can see the Kriging weights $\bw_i = (w_{i1},...,w_{im})^T$ are deterministic functions of $\bX_i = \{\bh_{i1},...,\bh_{iw_i}, r, \nu\}\equiv(X_{i1},...,X_{ip})$ where $\bh_{ij} = d_{ij}/\phi$ and $p = 2m+2$. Therefore, the Kriging weights can be written as:

\begin{equation} \label{eq:KW_W}
\begin{split}
\centering
    w_{ij} &= \tilde{w}_j(\bX_i), \\
    (w_{i1},...,w_{im}) &= \tilde{w}(\bX_i), \\
\end{split}
\end{equation}
where $\tilde{w}_j(\bX_i)$ is a function of the configuration of all $m$ conditioning-set locations and of $\boldsymbol{\theta}$. For a non-stationary covariance, it would be necessary to add $\bs_i$ to $\bX_i$ as the correlation function, and thus, the form of the Kriging weights would vary across space.

We approximate the deterministic function, $\tilde{w}(.)$ in \eqref{eq:KW_W} with an $m$-objective feed-forward neural network using the ReLU activation function: $\mbox{ReLU}(x) = x\cdot \mathbb{I}(x>0)$, where $\mathbb{I}$ is the indicator function. For $h \in \{1,\ldots,H\}$ hidden layers, weight matrix $\bW^{(.)}$ and bias vector $\bb^{(.)}$, we employ a multi-layer perceptron (MLP) structured as follows:
\begin{equation} \label{eq: FFNN_KW}
\small
\begin{split}
       \bx^{(h)} &= \mbox{ReLU}(\bW^{(h)}\bx^{(h-1)} + \bb^{(h)}),\,\\
    \tilde{w}(\bX_i) \approx \bx^{(H+1)} &= \bW^{(H+1)}\bx^{(H)} + \bb^{(H+1)},
\end{split}
\end{equation}
with $\bx^{(0)} = \bX_i$. We find through numerical experiments that an MLP with 2 hidden layers (i.e. H = 2) suffices for this study, however this can be easily adjusted by the user. 

The training data for the neural network is generated through $L$ simulation sets of varying total locations $n_l$, neighbor configurations $\bs_{(i)}$ and covariance parameter values $\boldsymbol{\theta}$. For each simulation in $1,\ldots,L$, location coordinates are independently sampled from a Uniform(0,1) distribution and $m$ neighbors are ordered using max-min distance ordering \citep{guinness2018permutation}, which has been shown to often be better in practice over alternate ordering methods for spatial datasets. The range and smoothness parameters are generated from design distribution $\phi \sim \mbox{Uniform}(0.005,0.11)$ and $\nu \sim \mbox{Uniform}(0.4,2.6)$ respectively. Network training is highly sensitive to the value of $r$, so $r$ is split into and uniformly sampled from 6 ranges covering (0.18, 1) with slight overlap: \{(0.18, 0.52], [0.38, 0.62], [0.58, 0.82], [0.78, 0.92], [0.88, 0.96], [0.94, 1)\}. We believe a true proportion of spatial variance less than 0.18 contains too much noise and a spatial model may not be appropriate. Collectively, the distributions for $\phi$, $\nu$, and $r$ that is used to generate data for training the MLP, are referred to as the design distribution and denoted by $\Pi$. The design distribution represents plausible ranges of covariance parameter values that real data are expected to have. Thus, for Kriging weight estimation, 6 different models are trained, each with $L$ simulations to cover the total range of $r$. Algorithm \ref{a:NN} describes this process.

\begin{algorithm}
    \caption{Training for A-BLINK.}\label{a:NN}
    \begin{algorithmic}[1]
    \Require Simulation sets, $L$; number of locations, $n_l, l=1,\ldots,L$ 
    \Require Conditioning set size, $m$; design distribution, $\boldsymbol{\Pi}$
    \For{$l \in \{1,...,L\}$}
      \State Simulate and order $n_l$ locations on $(0,1)^2$
      \State Draw $(\phi^{(l)}, r^{(l)},\nu^{(l)})\sim \boldsymbol{\Pi}$
      \State Calculate features $\bX_i^{(l)} = (\bh_{i1}^{(l)},...,\bh_{im}^{(l)}, r^{(l)}, \nu^{(l)})$, for $i=1,\ldots,n$
      \State Compute Kriging weights $\bw_l = (w_{l1},...,w_{lm})$ given $\bX_l$ using \eqref{eq:kriging_eqn}
    \EndFor
    \State Train neural network to learn a map from features to Kriging weights
    \end{algorithmic}
\end{algorithm}


Since the log-variance $w_{i0}$ is also a deterministic function of $\bX_i$ similar to \eqref{eq:KW_W},
\begin{equation} \label{eq:LV_W}
    w_{i0} = \tilde{w}_V(\bX_i),
\end{equation}

\noindent $\tilde{w}_V(.)$ is approximated similarly to the Kriging weights with an MLP. However, this network has only 1 output as opposed to $m$, and we find 1 hidden layer suffices to estimate with high accuracy. The inputs and data generation process are the same as the Kriging weight algorithm previously described, and have been omitted for brevity. The neural network will give a different variance for each spatial location. Adding the variance models together with the Kriging weights, we have 12 pre-trained networks for use in likelihood evaluation.

We will now briefly describe the network architectures and training process for the $m = 30$ neighbor models. The training data is simulated from $L=200$ sets of locations and parameter values, totaling around 2,000,000 training observations for each model. The Kriging weight MLPs have two hidden layers with 160 and 120 nodes, respectively. The log-variance MLPs have a single hidden layer with 40 nodes. All models were written and trained in torch for R \citep{torchR} using the default Adam \citep{kingma2014adam} optimizer with learning rate 0.001 and batch size of 500 for 25 epochs. The models can be trained in parallel, reducing the pre-training time cost.


\subsection{Likelihood evaluation}

The pre-trained networks amortize the cost of evaluating \eqref{eq:kriging_eqn} during MCMC or maximum likelihood estimation. As the networks are pre-trained, the evaluation of $\tilde{w}(\bX_i)$ and $\tilde{w}_V(\bX_i)$ are simply parallelizable matrix multiplications passed through the ReLU function, providing the amortization aspect of the algorithm.

The frequentist variant of A-BLINK is used for initial estimates and the estimate of $r$ is used to initially select which neural networks to use during MCMC sampling. A logit transform, $r_L = \mbox{logit}(r)$, where $\mbox{logit}(x) = \frac{1}{1+e^{-x}}$, such that it's value is no longer restricted to $[0,1]$. The following priors are then placed on the components of $\boldsymbol{\theta}$:
\begin{equation}\label{eq: Theta_Priors}
\begin{split}
    \pi(\phi) &\sim \text{Gamma}(p_1, p_2)\ , \\
    \pi(\nu) &\sim \text{LogNormal}(v_1, v_2)\ , \\
    \pi(r_L) &\sim \text{Normal}(r_1, r_2).
\end{split}
\end{equation}
Metropolis-Hastings \citep{robert2004metropolis} is used to update $\{\phi, \nu\}$ where the two are each sampled from a log-normal candidate distribution and $r_L$ is updated with Metropolis sampling with a normal candidate distribution. The neural networks replace evaluating \eqref{eq:kriging_eqn} during posterior calculation in \eqref{eq:Vecchia_Post} at each iteration of the MCMC, leading to greater amortization advantages.

\section{Simulation Study} \label{s:Simulation}

We tested A-BLINK against NNGP implemented in the spNNGP \citep{finley2020spnngp} package for various parameter configurations to show computational speedups gained through amortization. To our knowledge, NNGP is the most utilized state-of-the-art methodology in scalable Bayesian GP learning. We report MSE of the MCMC posterior draws, credible interval coverage for the parameters and Effective sample size (ESS) \citep{heidelberger1981spectral} per minute for timing comparison. 

The spatial locations are generated randomly on a 0-1 grid; the total number of training locations are drawn randomly from a Uniform(5000, 15000) distribution and the number of test locations drawn randomly from a Uniform(500, 1000) distribution. This centers the number of observed locations to be around 10,000, which is what the pre-trained neural network data was centered around. The GP data, $\bZ(\bs)$, is generated from 3 combinations of $\phi, \nu, r$, each with 100 independent simulations. For computational feasibility, these GP simulations are Vecchia approximated with $m = 80$ neighbors for each site. We also set the marginal mean and variance, $\mu$ and $\sigma^2$, equal to 0 and 1 respectively. Thus, the generated GP represents the standardized spatially correlated residuals in a regression analysis. Three parameter combinations were tested: $\boldsymbol{\theta}_1 = \{\phi = 0.01, \nu = 1.0, r = 0.4\}$,  $\boldsymbol{\theta}_2 = \{ \phi = 0.05, \nu = 2.0, r = 0.75\}$, and $\boldsymbol{\theta}_3 = \{\phi = 0.1, \nu = 1.5, r = 0.9\}$. This tests the methods over a wide variety of spatial range, smoothness and proportion of variance. 

The size of the conditioning set is set to be $m = 30$. The neighbors were selected using max-min ordering. For A-BLINK, we take 12,000 MCMC draws with 2,000 burn-in and the Metropolis-Hastings turning parameter set at 0.1 for all parameters. For NNGP, we take 8,000 draws with 4,000 burn-in and the tuning parameter set at 0.1. For A-BLINK, the following priors were used for the covariance parameters to assume a reasonable range of values:
\begin{equation}\label{eq: Sim_Priors}
\begin{split}
    \pi(\phi) &\sim \text{Gamma}(1.5, 30)\ , \\
    \pi(\nu) &\sim \text{LogNormal}(0.5, 0.5)\ , \\
    \pi(r_L) &\sim \text{Normal}(0, 1).
\end{split}
\end{equation}
For NNGP, the priors were:

\begin{equation}\label{eq: NNGP_Priors}
\begin{split}
    \pi(\phi) &\sim \text{Unif}(0.005, 0.15)\ , \\
    \pi(\nu) &\sim \text{Unif}(0.5, 2.5)\ , \\
    \pi(\tau^2) &\sim \text{InvGamma} (1, 0.2).
\end{split}
\end{equation}
\noindent NNGP estimates $r$ through the transform $r = \frac{1}{1+\tau^2}$. The initial values for NNGP were set to be the true parameter values for computational stability. These represent weakly informative priors that place reasonable ranges on each of the parameters. NNGP requires Uniform priors on $\phi$ and $\nu$ and Inverse Gamma prior on $\tau^2$, hence the choices of prior distribution made. The prior 95\% credible intervals for each of the parameters is shown in Table \ref{tab: PriorCI}.

\begin{table}[ht]
\footnotesize
 \centering
 \caption{Prior 95\% CI on A-BLINK and NNGP parameters.} 
 \label{tab: PriorCI}
 \resizebox{7cm}{!}{
 \begin{tabular}{|l|c|c|}
   \hline \textbf{Algorithm}  & Parameter & 95\% CI \\  \hline
   &                    $\phi$ & (0.00, 0.16) \\ 
   \textbf{A-BLINK} &   $\nu$ & (0.600, 4.42) \\
   &                    $r$     &  (0.11, 0.86)\\
    \hline
   &               $\phi$ & (0.001, 0.15) \\ 
   \textbf{NNGP} &   $\nu$ & (0.55, 2.44) \\
   &                   $r$       & (0.12, 0.95) \\
  \hline
 \end{tabular}
 }
 \end{table}

\subsection{Simulation Results}

\begin{figure*}
    \centering
    \includegraphics[width=\linewidth, scale = 1]{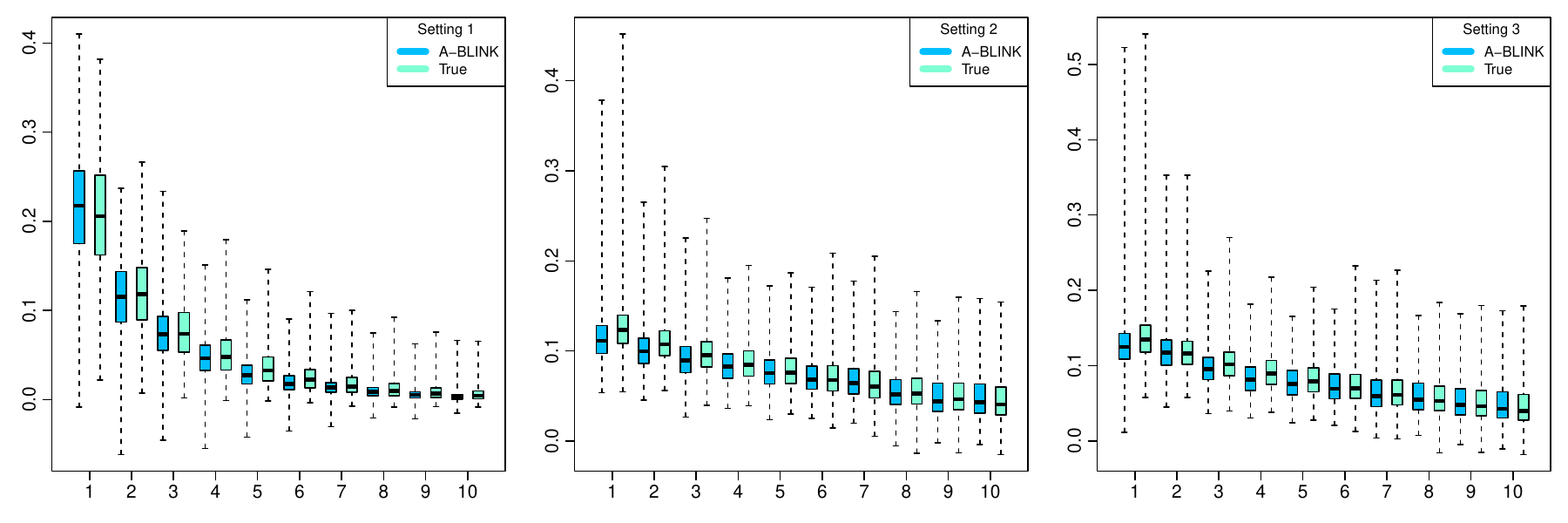}
    \caption{Boxplots of the first 10 true and A-BLINK predicted kriging weights for the three simulation settings, each based on a single dataset with the same locations.}
    \label{fig:weights_sims}
 \end{figure*}

Figure \ref{fig:weights_sims} shows a comparison of the true and A-BLINK predicted kriging weights for all 3 settings, based on a single field of 7,625 locations. Only the first 10 kriging weights are plotted for clarity of presentation. We see that the predicted and true kriging weights have very similar distributions. This was observed across all settings, indicating that the neural network is able to reliably predict the kriging weights. We note that when either the range or the proportion of spatial variance is large (or both), the squared correlation ($R^2$) values between the true and the predicted weights remain higher than $0.90$ for all neighbors. However, when the range or the proportion of variance is small (or both, as in $\boldsymbol{\theta}_1$,) the farther neighbors have much lower $R^2$. This is expected since the response has much weaker dependence on the farther-out neighbors, and they are not important variables to the neural network prediction in these situations. As we see next, this does not have any perceivable bearing on parameter estimation.

Table \ref{tab: Bayes_MSE} gives the MSE of the estimated covariance parameters in each of the 3 settings. A-BLINK drastically outperforms NNGP in MSE across all settings with very low error. We find that the results for NNGP are highly sensitive to the choice of prior and concentrate the prior around reasonable values for all 3 parameters for our study.

\begin{table}[ht]
 \centering
 \caption{Parameter estimation MSE for $\boldsymbol{\theta}_1,\boldsymbol{\theta}_2,\boldsymbol{\theta}_3$ between A-BLINK and NNGP. Results are computed over 100 simulations. Lower values are better. Standard errors are in parenthesis.} 
 \label{tab: Bayes_MSE}
 \resizebox{10cm}{!}{
 \begin{tabular}{|c|l|c|c|c|}
   \hline $\boldsymbol{\theta}$ & \textbf{Algorithm} & $\boldsymbol{\phi} \textbf{ Error}$ & $\boldsymbol{\nu} \textbf{ Error}$ & \textbf{$r$ Error} \\  \hline
   1 & \textbf{A-BLINK} & 0.000 (0.000) & 0.063 (0.007) & 0.001 (0.000) \\ 
 & \textbf{NNGP} & 0.009 (0.001) & 0.364 (0.041) & 0.014 (0.000) \\ 
  2 & \textbf{A-BLINK} & 0.000 (0.000) & 0.128 (0.015) & 0.000 (0.000) \\ 
 & \textbf{NNGP} & 0.008 (0.000) & 1.794 (0.079) & 0.001 (0.000) \\  
  3 & \textbf{A-BLINK} & 0.000 (0.000) & 0.015 (0.002) & 0.000 (0.000) \\ 
 & \textbf{NNGP} & 0.002 (0.000) & 0.991 (0.007) & 0.000 (0.000) \\ 
    \hline
 \end{tabular}
 }
 \end{table}

Table \ref{tab: Bayes_Cov} represents the coverage of the 95\% credible intervals for A-BLINK and NNGP. We note that A-BLINK's coverage of proportion of variance is low for setting 3. In setting 3, the true proportion of 0.95 is at the intersection of two neural network models, causing oscillation between both models during likelihood calculation and impacting sampling of the proportion parameter. Despite coverage being lower than desired, from the prior MSE table, the posterior means are still centered around the true values for all A-BLINK estimated parameters. Additionally, the coverage from NNGP is very poor in comparison to A-BLINK, with credible intervals often failing to capture the true value. We again attribute this to extreme prior sensitivity.

 \begin{table}[ht]
 \centering
 \caption{Parameter estimation coverage for $\boldsymbol{\theta}_1,\boldsymbol{\theta}_2,\boldsymbol{\theta}_3$ between A-BLINK and NNGP. Results are computed over 100 simulations. Higher values are better. Standard errors are in parenthesis.} 
 \label{tab: Bayes_Cov}
 \resizebox{10cm}{!}{
 \begin{tabular}{|c|l|c|c|c|}
   \hline $\boldsymbol{\theta}$ & \textbf{Algorithm} & $\boldsymbol{\phi} \textbf{ Coverage}$ & $\boldsymbol{\nu} \textbf{ Coverage}$ & \textbf{$r$ Coverage} \\  \hline
 1 & \textbf{A-BLINK} & 0.880 (0.032) & 0.820 (0.038) & 0.920 (0.027) \\ 
 & \textbf{NNGP} & 0.180 (0.038) & 0.420 (0.049) & 0.000 (0.000) \\ 
 2 & \textbf{A-BLINK} & 0.850 (0.036) & 0.840 (0.037) & 0.900 (0.030) \\ 
 & \textbf{NNGP} & 0.170 (0.038) & 0.180 (0.038) & 0.120 (0.032) \\ 
 3 & \textbf{A-BLINK} & 0.930 (0.026) & 0.980 (0.014) & 0.680 (0.047) \\ 
 & \textbf{NNGP} & 0.020 (0.014) & 0.010 (0.010) & 0.510 (0.050) \\  
    \hline
 \end{tabular}
 }
 \end{table}

Table \ref{tab: Bayes_ESS} summarizes the ESS per minute (ESS/min) for A-BLINK and NNGP, showing the sampling efficiency for each method. A-BLINK is drastically more efficient than NNGP with ESS/min being at least 150 times higher for the range and smoothness parameters and proportion being at least 7.5 times higher. This shows A-BLINK is able to adapt to the posterior distribution and converge faster than NNGP with both using Metropolis-Hastings samplers and similar prior parameter ranges.

\begin{table}[ht]
 \centering
 \caption{Parameter effective sample size per minute for $\boldsymbol{\theta}_1,\boldsymbol{\theta}_2,\boldsymbol{\theta}_3$ between A-BLINK and NNGP. Results are computed over 100 simulations. Higher values are better. Standard errors are in parenthesis.} 
 \label{tab: Bayes_ESS}
 \resizebox{10cm}{!}{
 \begin{tabular}{|c|l|c|c|c|}
   \hline $\boldsymbol{\theta}$ & \textbf{Algorithm} & $\boldsymbol{\phi} \textbf{ ESS/min}$ & $\boldsymbol{\nu} \textbf{ ESS/min}$ & \textbf{$r$ ESS/min} \\  \hline
 1 & \textbf{A-BLINK} & 12.908 (1.125) & 11.640 (1.012) & 25.269 (2.434) \\ 
 & \textbf{NNGP} & 0.033 (0.002) & 0.037 (0.003) & 3.344 (0.165) \\ 
 2 & \textbf{A-BLINK} & 7.368 (0.503) & 6.986 (0.499) & 70.574 (6.465) \\ 
 & \textbf{NNGP} & 0.044 (0.003) & 0.047 (0.004) & 2.929 (0.210) \\ 
 3 & \textbf{A-BLINK} & 15.162 (0.637) & 14.042 (0.537) & 32.913 (1.191) \\ 
 & \textbf{NNGP} & 0.046 (0.003) & 0.047 (0.003) & 3.586 (0.174) \\ 
    \hline
 \end{tabular}
 }
 \end{table}

Finally, Table \ref{tab: Bayes_Pred} shows the prediction MSE for the test locations for both methods. For A-BLINK, the parameter values every 50 iterations are used to make predictions using the GpGp \citep{guinness2021gaussian} package while NNGP uses its own predict function. A-BLINK has around 10\% lower MSE for the low proportion of spatial variance setting 1 with slightly lower MSE for settings 2 and 3, showing predictive performance doesn't suffer despite the additional approximation from the neural network weights and variance. Simulations with very tight priors centered around the true parameter values for NNGP were also tested. While parameter MSE was very low, as expected, coverage was still lacking and the ESS/min did not improve. Thus, these results are not reported.

\begin{table}[ht]
 \centering
 \caption{Prediction MSE for $\boldsymbol{\theta}_1,\boldsymbol{\theta}_2,\boldsymbol{\theta}_3$ between A-BLINK and NNGP. Results are computed over 100 simulations. Lower values are better. Standard errors are in parenthesis.} 
 \label{tab: Bayes_Pred}
 \resizebox{10cm}{!}{
 \begin{tabular}{|l|c|c|c|}
   \hline \textbf{Algorithm} & $\boldsymbol{\theta}_1$ & $\boldsymbol{\theta}_2$ & $\boldsymbol{\theta}_3$ \\  \hline
 \textbf{A-BLINK} & 0.803 (0.005) & 0.266 (0.001) & 0.107 (0.001) \\ 
 \textbf{NNGP} & 0.890 (0.006) & 0.270 (0.002) & 0.109 (0.001) \\ 
    \hline
 \end{tabular}
 }
 \end{table}

\section{Analysis of mean temperature data for the US} \label{s:RealData}

We demonstrate our methodology on mean temperature data from 1991--2020 US Climate Normals dataset. The product, provided by the National Centers for Environmental Information\footnote{\href{https://www.ncei.noaa.gov/products/land-based-station/us-climate-normals}{US Climate Normals}}, measures decadal changes in meteorological variables across the US. It is usually generated once every 10 years, with the 1991--2020 dataset being the latest iteration. Our dataset consists of 30-year mean temperatures in Fahrenheit at 7306 weather stations across the contiguous US, indexed by latitude and longitude, as well as station elevation.
\begin{figure}[t]
    \centering
    \includegraphics[width= 0.9\linewidth, scale = 1.5]{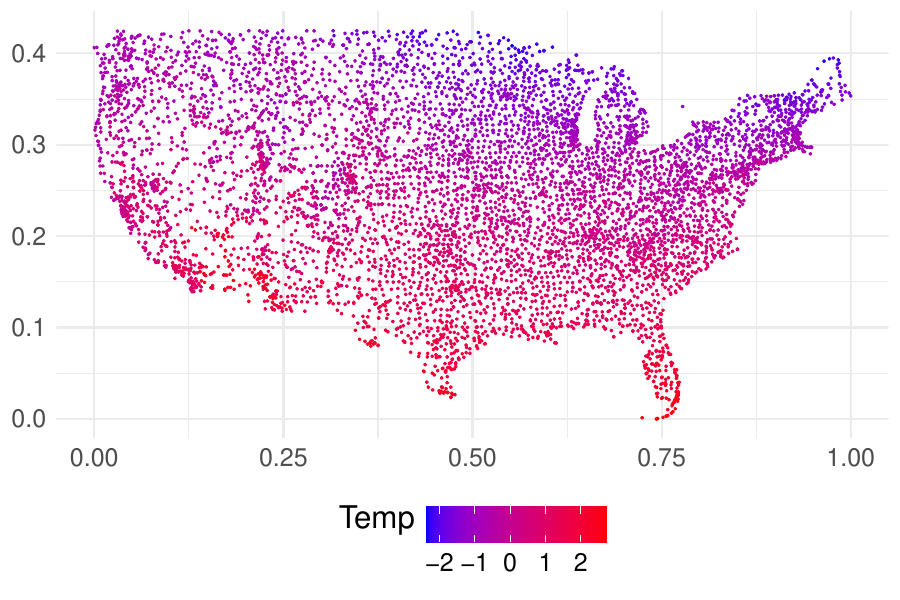}
    \caption{Standardized 30-year mean temperatures based on the 1991--2020 US climate normals dataset. Plotted using coordinates translated to the unit square.}
    \label{fig:ClimNorms}
 \end{figure}

\begin{figure*}
    \centering
    \includegraphics[width=0.9\linewidth, scale = 1.5]{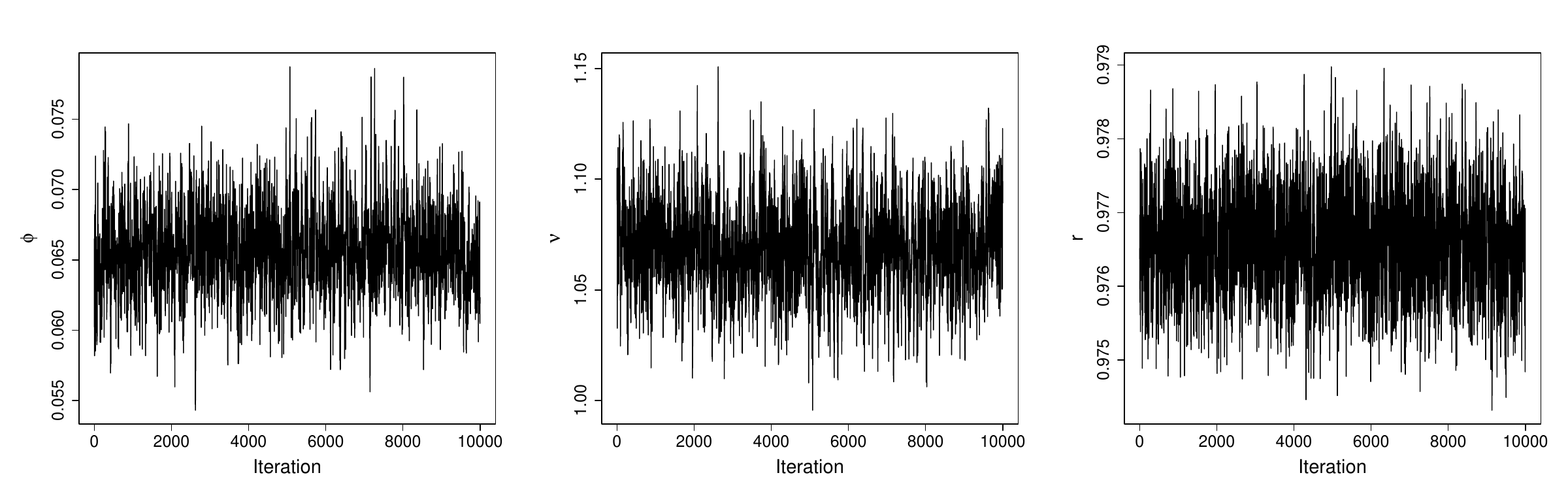}
    \caption{Trace plots from A-BLINK for the posterior distribution of the covariance parameters for the mean temperature dataset.}
    \label{fig:MCMC_Clim}
 \end{figure*}

To standardize the data, temperature is first regressed onto elevation to estimate the marginal mean and variance. We subtract the fitted temperature from the truth and divide by the residual standard deviation to create our standardized temperature response. The response can be interpreted as the standardized residuals of the regression and is visualized in Figure \ref{fig:ClimNorms}. The data shows a pronounced spatial pattern, with lower temperatures as we go further north. From the standardized mean temperature, we create a training and test set using a 90/10\% train/test split based on the locations. Both models are fit with 11,000 iterations with the same priors used in the simulation section. The first 1,000 and 5,500 iterations are discarded as burn in for A-BLINK and NNGP respectively. Trace plots for the posterior distribution of the covariance parameters based on our approach are included in Figure \ref{fig:MCMC_Clim}, showing convergence for all three parameters. We compare predictive power of both methods through test set MSE and $R^2$ between predicted and true mean temperature values. Estimation efficiency is compared through parameter ESS/min from the training set.

\begin{table}[ht]
 \centering
 \caption{Climate Normals analysis results for A-BLINK and NNGP. Prediction MSE and $R^2$ is first shown, followed by parameter posterior mean, 95\% CI and ESS/min.} 
 \label{tab: Clim_Norms}
 \resizebox{12cm}{!}{
 \begin{tabular}{|l|c|c|c|c|}
   \hline \textbf{Algorithm} & MSE ($R^2$) & Parameter & Estimate (95\% CI) & ESS/min \\  \hline
   &                                  & $\phi$ & 0.07 (0.06, 0.07) & 16.18 \\ 
   \textbf{A-BLINK} & 0.029 (0.973)   & $\nu$ & 1.07 (1.03, 1.11)  & 15.21\\
   &                                  & $r$      & 0.98 (0.98, 0.98) & 37.68\\
    \hline
   &  & $\phi$ & 0.15 (0.15, 0.15) & 0.06 \\ 
   \textbf{NNGP} & 0.034 (0.968)    &  $\nu$ & 0.50 (0.50, 0.50) & 0.09\\
   &                         & $r$       & 0.96 (0.96, 0.96) & 11.41 \\
  \hline
 \end{tabular}
 }
 \end{table}

As we can see in Table \ref{tab: Clim_Norms}, A-BLINK has slightly better predictive power, although both have high test set $R^2$ at around 97\%. In parameter estimation, NNGP fails to explore the full posterior space, seeming to get stuck at the prior edges while A-BLINK is able to explore a wider posterior space. This results in drastically higher ESS/min with $\phi$ being around 270 times more efficient and $\nu$ being around 170 times more efficient. We found NNGP to be very sensitive to the choice of prior. Using these weakly informative priors was not sufficient for NNGP, whereas A-BLINK was much more robust. 

\section{Discussion}

Recent advances in amortized learning have provided pathways to statistical inference on large spatial datasets, bypassing expensive likelihood evaluations that were previously computationally infeasible. This has especially allowed for both significantly faster parameter estimation and response prediction for GPs. However, the use of amortized learning for GP inference in a Bayesian setting has yet to be well explored. We introduce the A-BLINK framework that employs the Vecchia decomposition to approximate the GP likelihood, and uses pre-trained neural networks to learn the Kriging weights that are needed to evaluate the likelihood. This bypasses the need for matrix inversions and allows it to be embedded in Bayesian hierarchical models for full posterior inference. The computational cost of covariance matrix inversions to calculate Kriging weights and variance is amortized with our method, significantly reducing likelihood evaluation time. The framework is easily adaptable to user-specified parameter ranges and arbitrary irregular spatial fields with over 10,000 locations. From the simulation study, we see that A-BLINK is robust to noisy datasets with low proportion of spatial variance. Our approach is faster and more scalable than fully Bayesian methods which inevitably require matrix inversions, and is more flexible than competing machine learning methods which are usually designed for gridded data products.

This work focuses on estimation and inference of covariance parameters for GPs with standard normal marginals. A fixed regression is often used to model the marginal mean process in GPs. Assuming a regression for the mean and marginal variance not equal to 1, we derived posterior distributions for these parameters within the A-BLINK framework and include these in the supplementary material. Future work will involve estimating marginal means and variances within the hierarchical Bayesian framework, following the approach of \citet{majumder2024modeling} and \citep{majumder2023deep}.

Our choice of using an MLP architecture was influenced by its simplicity, since prediction using the network involves only a series of matrix multiplications and additions. Similarly, the tunable hyperparameters of the network (configuration of nodes, learning rates, batch sizes etc.) was based on rules of thumb which seemed to provide a reasonable balance of speed and accuracy. These are aspects we intend to test more rigorously in the future, since we anticipate that these can be further optimized to provide a better blend of predictive accuracy and computational overhead. This would be particularly important as number of locations in our datasets keep increasing. Finally, we would like to combine the two separate networks for kriging weights and variance into one network with two output heads, one for the weights and one for the variance. As the weights and variance are connected in the exact weight calculations, this connection should be exploited during network training. 

\section{Computing Details}

The simulation study results were run on a compute cluster fitted with Intel Xeon processors. The climate norms analysis and neural network pre-training were run on a Macbook Pro with an M1 Max processor.

\begin{singlespace}
 \bibliographystyle{rss}
	\bibliography{ref}
\end{singlespace}

\newpage

\appendix
\section{Appendix}

\subsection{Frequentist Results}

\subsubsection{Frequentist computation algorithm}

Algorithm \ref{a:Par_est} details the steps for frequentist parameter estimation. A variogram is first fit onto the data to get initial estimates of $\boldsymbol{\theta} = \{\phi, \nu, r\}$ denoted as $\{\phi_v, \nu_v, r_v\}$. The pre-trained Kriging and log-variance networks are then loaded according to which of 6 proportion of spatial variance ranges that $r_v$ falls under. The final maximum likelihood estimates, $\{\phi_N, \nu_N, r_N\}$, are found utilizing L-BFGS-B for constrained optimization \citep{byrd1995limited}. We constrain the parameter bounds to the pre-training bounds of (0.005, 0.12) for $\phi$, (0.3, 2.7) for $\nu$ and (0.18, 0.99) for $r$. For predictions, $r_N$ is used with the log-variance network to estimate the parameterization of the spatial variance and nugget needed by the GpGp predict function.

\setcounter{algorithm}{1}
\begin{algorithm}[H]
    \caption{Algorithm for frequentist spatial parameter estimation and predictions}\label{a:Par_est}
    \begin{algorithmic}[1]
    \Require Observed neighbor set $\boldsymbol{X}_L$; Prediction set $\boldsymbol{X}_T$
    \State Find variogram initial estimates $\phi_v, \nu_v, r_v$
    \State Load in appropriate Kriging and log-variance networks
    \State Find $\phi_N, \nu_N, r_N$ from optimizing multivariate normal likelihood with initial values $\phi_v, \nu_v, r_v$
    \State Use $r_N$ with log-variance network to estimate $r_N\sigma^2_N$ and $(1-r_N)\sigma^2_N$
    \State Use GpGp prediction function with $r_N\sigma^2_N, \phi_N, \nu_N, (1-r_N)\sigma^2_N$
    \end{algorithmic}
\end{algorithm}

\subsubsection{Frequentist Simulation Study}

We tested frequentist A-BLINK against GpGp for the same three parameter configurations to show computational speedups gained through amortization. To our knowledge, GpGp is the most efficient and most utilized package for frequentist Vecchia approximated GP learning. We report MSE of the parameter estimates, the total algorithm runtime and the prediction set MSE. 

The spatial locations are generated randomly on a 0-1 grid; the total number of training locations are drawn randomly from a Uniform(5000, 15000) distribution and the number of test locations drawn randomly from a Uniform(500, 1000) distribution. This centers the number of observed locations to be around 10,000, which is what the pre-trained neural network data was centered around. The GP data, $\boldsymbol{Z}(\boldsymbol{s})$, is generated from 3 combinations of $\phi, \nu, r$, each with 100 independent simulations. For computational feasibility, these GP simulations are Vecchia approximated with $m = 80$ neighbors for each site. We also set the marginal mean and variance, $\mu$ and $\sigma^2$, equal to 0 and 1 respectively. Thus, the generated GP represents the standardized spatially correlated residuals in a regression analysis. Three parameter combinations were tested: $\boldsymbol{\theta}_1 = \{\phi = 0.01, \nu = 1.0, r = 0.4\}$,  $\boldsymbol{\theta}_2 = \{ \phi = 0.05, \nu = 2.0, r = 0.75\}$, and $\boldsymbol{\theta}_3 = \{\phi = 0.1, \nu = 1.5, r = 0.9\}$. This tests the methods over a wide variety of spatial range, smoothness and proportion of variance. 

The size of the conditioning set is set to be $m = 30$. The neighbors were selected using max-min ordering. For A-BLINK, we set the convergence tolerance at 1e-7. For GpGp, we set the number of preconditioning neighbors as 10 and the convergence tolerance as 1e-4. These are the default values GpGp uses.

\subsubsection{Results}
Table \ref{tab: Combined_Freq_Set} shows the comparison with GpGp for frequentist estimation. While neither algorithm is consistently better for parameter estimation across all 3 settings, A-BLINK is notably faster. When the true proportion of spatial variance is high ($\boldsymbol{\theta}_2, \boldsymbol{\theta}_3$), it is around 2-3 times faster than GpGp. When the true proportion of spatial variance is low ($\boldsymbol{\theta}_1$), A-BLINK converges nearly 5 times faster than GpGp. We see that GpGp struggles with parameter estimation accuracy when the true proportion of spatial variance is low, which also affects its convergence time. We suspect that the added noise in the data is the cause for this. Although GpGp is slightly better in prediction MSE, this can be expected as it is only using one level of approximation in the likelihood while we have two: the Vecchia approximation of the full likelihood and the neural network estimator of the Kriging weights and log-variance of the Vecchia approximated likelihood. We also note that the difference in predictive performance is negligible compared to the added computation cost of GpGp.

\begin{table}[ht]
\centering
\caption{Frequentist parameter estimation and prediction MSE for $\boldsymbol{\theta}_1,\boldsymbol{\theta}_2,\boldsymbol{\theta}_3$ between A-BLINK and GpGp. Results are computed over 100 simulations. Lower values are better. Standard errors are in parenthesis.} 
\label{tab: Combined_Freq_Set}
\resizebox{15cm}{!}{

\begin{tabular}{|c|l|c|c|c|c|c|}
  \hline $\boldsymbol{\theta}$ & \textbf{Algorithm} & $\boldsymbol{\phi} \textbf{ Error}$ & $\boldsymbol{\nu} \textbf{ Error}$ & \textbf{r Error} & \textbf{Time (s)} & \textbf{Pred MSE} \\ 
  \hline
  1 & \textbf{A-BLINK} & 0.000 (0.000) & 0.127 (0.017) & 0.003 (0.000) & 7.445 (0.232) & 0.804 (0.005) \\ 
  & \textbf{GpGp} & 0.000 (0.000) & 0.217 (0.047) & 0.055 (0.003) & 34.450 (1.003) & 0.800 (0.005) \\
  \hline
  2 & \textbf{A-BLINK} & 0.000 (0.000) & 0.285 (0.023) & 0.000 (0.000) & 6.779 (0.220) & 0.266 (0.001) \\
  & \textbf{GpGp} &  0.000 (0.000) & 0.142 (0.018) & 0.003 (0.000) & 19.213 (0.556) & 0.266 (0.001) \\
  \hline
  3 & \textbf{A-BLINK} & 0.000 (0.000) & 0.138 (0.023) & 0.001 (0.000) & 6.949 (0.189) & 0.107 (0.001) \\ 
  & \textbf{GpGp} & 0.001 (0.000) & 0.043 (0.007) & 0.000 (0.000) & 15.420 (0.455) & 0.107 (0.001) \\
  \hline
\end{tabular}
}
\end{table}

\subsection{Marginal Posterior Derivations}

\subsubsection{Posterior of $\beta$}

Given observed data $Z_1,...,Z_n$, covariate matrix $\boldsymbol{X} \in \mathbb{R}^{n \times p}$ and parameter set $\Theta = \{\boldsymbol{\beta}, \sigma^2, \phi, \nu, r\}$ with distribution:

$$Z_i|\boldsymbol{Z}_{(i)},\Theta \sim \mbox{Normal}\left(\boldsymbol{X}_i\boldsymbol{\beta}+\sum_{j=1}^{m}w_{ij}(Z_{\dot{\iota}_{j}}-\boldsymbol{X}_j\boldsymbol{\beta}),\sigma^2\exp(w_{i0})\right),$$

\noindent
and prior on $\boldsymbol{\beta}$: 

$$\pi(\boldsymbol{\beta}) \sim N(0, \sigma^2_\beta I_p),$$ 

let $Z_i - \sum_{j=1}^m w_{ij}Z_{i_j} = Z_i^*$ and $(\boldsymbol{X}_i - \sum_{j=1}^m w_{ij} \boldsymbol{X}_j) = \boldsymbol{X}_i^*$.
\vspace{\baselineskip}

The posterior of $\boldsymbol{\beta}$, $p(\boldsymbol{\beta})$ is calculated as follows:

\begin{equation}
    p(\boldsymbol{\beta}) \propto  \exp(\frac{-1}{2}(Z^* - \boldsymbol{X}^*\boldsymbol{\beta})^T(\frac{1}{\sigma^2\exp(w_{i0})}I_{n})(Z^* - \boldsymbol{X}^*\boldsymbol{\beta})) \\ \exp(-\frac{1}{2}(\boldsymbol{\beta}^T\frac{1}{\sigma^2_\beta}\boldsymbol{\beta}))
\end{equation}

\noindent We can write this as

\begin{equation}
    \exp(-\frac{1}{2}(-2Z^{*T}\frac{1}{\sigma^2\exp(w_0)}\boldsymbol{X}^*\boldsymbol{\beta} + \boldsymbol{\beta}^T\boldsymbol{X}^{*T}\frac{1}{\sigma^2\exp(w_0)}\boldsymbol{X}^*\boldsymbol{\beta} + \boldsymbol{\beta}^T\frac{1}{\sigma^2_\beta}\boldsymbol{\beta})
\end{equation}

which equals 

\begin{equation}
    \exp(-\frac{1}{2}(-2Z^{*T}\frac{1}{\sigma^2\exp(w_0)}\boldsymbol{X}^*\boldsymbol{\beta} + \boldsymbol{\beta}^T(\boldsymbol{X}^{*T}\frac{1}{\sigma^2\exp(w_0)}\boldsymbol{X}^* + \frac{1}{\sigma^2_\beta})\boldsymbol{\beta})
\end{equation}

\noindent Let $A = Z^{*T}\frac{1}{\sigma^2\exp(w_0)}\boldsymbol{X}^*$ and $B = \boldsymbol{X}^{*T}\frac{1}{\sigma^2\exp(w_0)}\boldsymbol{X}^* + \frac{1}{\sigma^2_\beta}$

\noindent
Thus:
\begin{equation}
    p(\boldsymbol{\beta}) \sim N\left( (\boldsymbol{X}^{*T}\frac{1}{\sigma^2\exp(w_0)}\boldsymbol{X}^* + \frac{1}{\sigma^2_\beta})^{-1}(\boldsymbol{X}^{*T}\frac{1}{\sigma^2\exp(w_0)}Z), (\boldsymbol{X}^{*T}\frac{1}{\sigma^2\exp(w_0)}\boldsymbol{X}^* + \frac{1}{\sigma^2_\beta})^{-1} \right)
\end{equation}

\subsubsection{Posterior of $\sigma^2$}

Given observed data $Z_1,...,Z_n$, covariate matrix $\boldsymbol{X} \in \mathbb{R}^{n \times p}$ and parameter set $\Theta = \{\boldsymbol{\beta}, \sigma^2, \phi, \nu, r\}$ with distribution:

$$Z_i|\boldsymbol{Z}_{(i)},\Theta \sim \mbox{Normal}\left(\boldsymbol{X}_i\boldsymbol{\beta}+\sum_{j=1}^{m}w_{ij}(Z_{\dot{\iota}_{j}}-\boldsymbol{X}_j\boldsymbol{\beta}),\sigma^2\exp(w_{i0})\right)$$

\noindent
and given prior on $\sigma^2$: 

$$\pi(\boldsymbol{\beta}) \sim IG(a, b),$$

\noindent
let $Z_i - \sum_{j=1}^m w_{ij}Z_{i_j} = Z_i^*$ and $(\boldsymbol{X}_i - \sum_{j=1}^m w_{ij} \boldsymbol{X}_j) = \boldsymbol{X}_i^*$

\vspace{\baselineskip}

The posterior of $\sigma^2$, $p(\sigma^2)$ is calculated as follows:

\begin{equation}
    p(\sigma^2) \propto  (\sigma^2)^{-n/2} \exp(\frac{-1}{2\sigma^2} \sum \frac{1}{\exp(w_{i0})}(Z_i^* - \boldsymbol{X}_i^*\boldsymbol{\beta})^2))(\sigma^2)^{-a-1}\exp(\frac{-b}{\sigma^2})
\end{equation}

which equals

\begin{equation}
    p(\sigma^2) \sim  (\sigma^2)^{-(n/2 + a) - 1} \exp(\frac{-1}{\sigma^2}(\sum \frac{1}{2\exp(w_{i0})}(Z_i^* - \boldsymbol{X}_i^*\boldsymbol{\beta})^2 + b))
\end{equation}

\noindent
Thus:

\begin{equation}
    p(\sigma^2) \sim  IG((n/2 + a), \sum \frac{1}{2\exp(w_{i0})}(Z_i^* - \boldsymbol{X}_i^*\boldsymbol{\beta})^2 + b) 
\end{equation}

\end{document}

%% file: commands.tex
\newcommand{\bSigma}{ \mbox{\boldmath $\Sigma$}}

\newcommand{\bx}{ \mbox{\bf x}}
\newcommand{\bX}{ \mbox{\bf X}}

\newcommand{\bZ}{ \mbox{\bf Z}}

\newcommand{\bh}{ \mbox{\bf h}}

\newcommand{\bs}{ \mbox{\bf s}}
\newcommand{\bb}{ \mbox{\bf b}}

\newcommand{\bW}{ \mbox{\bf W}}
\newcommand{\bw}{ \mbox{\bf w}}

\newcommand{\calD}{{\cal D}}

\newcommand{\calN}{{\cal N}}

\newcommand{\Matern}{ \mbox{Mat$\acute{\mbox{e}}$rn}}

\newcommand{\beq}{ \begin{equation}}
\newcommand{\eeq}{ \end{equation}}
\newcommand{\beqn}{ \begin{eqnarray}}
\newcommand{\eeqn}{ \end{eqnarray}}

\usepackage{caption}